\documentclass[lettersize,journal]{IEEEtran}
\usepackage{amsmath,amsfonts}
\usepackage{algorithmic}
\usepackage{array}
\usepackage[caption=false,font=normalsize,labelfont=sf,textfont=sf]{subfig}
\usepackage{textcomp}
\usepackage{stfloats}
\usepackage{url}
\usepackage{verbatim}
\usepackage{graphicx}
\usepackage{xcolor}
\usepackage{cite}
\usepackage{amsmath,amssymb,amsfonts}
\usepackage{algorithmic}
\usepackage{graphicx}
\usepackage{textcomp}
\usepackage{multirow}
\usepackage{multicol}
\usepackage{makecell}
\usepackage[ruled]{algorithm2e}
\usepackage[utf8]{inputenc} 
\usepackage[T1]{fontenc}    
\usepackage{url}            
\usepackage{booktabs}       
\usepackage{nicefrac}       
\usepackage{microtype}      
\usepackage{colortbl}
\usepackage[backref]{hyperref}

\usepackage{amsmath}
\usepackage{amsfonts}
\usepackage{amssymb}
\usepackage{bm}

\def\BibTeX{{\rm B\kern-.05em{\sc i\kern-.025em b}\kern-.08em
    T\kern-.1667em\lower.7ex\hbox{E}\kern-.125emX}}
\usepackage{balance}
\begin{document}
\title{LViT: Language meets Vision Transformer in Medical Image Segmentation}
\author{Zihan Li, Yunxiang Li, Qingde Li, Puyang Wang, Dazhou Guo, Le Lu, \IEEEmembership{Fellow, IEEE}, \\Dakai Jin, \IEEEmembership{Member, IEEE}, You Zhang, Qingqi Hong, \IEEEmembership{Member, IEEE}
\thanks{Zihan Li is with Xiamen University and the Department of Computer Science, University of Illinois at Urbana-Champaign, Urbana, IL 61801, USA (e-mail: \href{zl111@illinois.edu}{zl111@illinois.edu}).}
\thanks{Yunxiang Li and You Zhang are with the Department of Radiation Oncology, UT Southwestern Medical Center, Dallas, TX 75235, USA.}
\thanks{Qingde Li is with the Department of Computer Science, University of Hull, Hull, HU6 7RX, UK.}
\thanks{Puyang Wang is with DAMO Academy, Alibaba Group, Hangzhou 310024, China.}
\thanks{Dazhou Guo, Le Lu, and Dakai Jin are with DAMO Academy, Alibaba Group, New York, NY 10014, USA.}
\thanks{Qingqi Hong is with Xiamen University, Xiamen 361005, China. (e-mail:
\href{hongqq@xmu.edu.cn}{hongqq@xmu.edu.cn}).}
\thanks{Corresponding author: Qingqi Hong}
}
\maketitle
\begin{abstract}
Deep learning has been widely used in medical image segmentation and other aspects. However, the performance of existing medical image segmentation models has been limited by the challenge of obtaining sufficient high-quality labeled data due to the prohibitive data annotation cost. To alleviate this limitation, we propose a new text-augmented medical image segmentation model LViT (Language meets Vision Transformer). In our LViT model, medical text annotation is incorporated to compensate for the quality deficiency in image data. In addition, the text information can guide to generate pseudo labels of improved quality in the semi-supervised learning. We also propose an Exponential Pseudo label Iteration mechanism (EPI) to help the Pixel-Level Attention Module (PLAM) preserve local image features in semi-supervised LViT setting. In our model, LV (Language-Vision) loss is designed to supervise the training of unlabeled images using text information directly. For evaluation, we construct three multimodal medical segmentation datasets (image + text) containing X-rays and CT images. Experimental results show that our proposed LViT has superior segmentation performance in both fully-supervised and semi-supervised setting. The code and datasets are available at {\href{https://github.com/HUANGLIZI/LViT}{https://github.com/HUANGLIZI/LViT}}.
\end{abstract}


\section{Introduction}
\label{sec:introduction}
\IEEEPARstart{M}{edical} image segmentation is one of the most critical tasks in medical image analysis. In clinical practice, accurate segmentation results are often achieved manually or semi-automatically. It remains a challenging task to extract the desired object accurately, especially when the target organ to be extracted is of high complexity in terms of tissue structures. Recent research shows that deep learning can be a promising approach for automatic medical image segmentation, as the knowledge of experts can be learned and extracted by using a certain deep learning method. A summary of existing solutions is shown in Figure 1(a):  (1) one shared encoder followed by two separate decoders \cite{58chen2019med3d}; (2) two separate encoders followed by one shared decode \cite{57wang2021boundary}; (3) two separate encoders followed by a modality interaction model \cite{59zhu2020lymph}. 
However, two inherent issues concerning the creation of high quality medical image datasets severely limit the application: one is the difficulty in obtaining high-quality images, and the other one is the high cost of data annotation \cite{1zhang2021self,2li2021dual}. These two issues have dramatically limited the performance improvement of medical image segmentation models. Since it is challenging to improve the quantity and quality of medical images themselves, it may be more feasible to use complementary and easy-to-access information to make up for the quality defects of medical images. Thus, we turn our attention to the written medical notes accompanied by medical images. It is well known that text data of medical records are usually generated along with the patients, so no extra cost is needed to access the corresponding text data. The medical text record data and the image data are naturally complementary to each other, so the text information can compensate for the quality deficiency in the medical image data.
On the other hand, expert segmentation annotation is often expensive and time-consuming, especially for new diseases like COVID-19, where high-quality annotations are even more difficult to obtain \cite{1zhang2021self, li2022tfcns, yu2022transfer}. In order to address the issue of under-annotated data, some approaches have gone beyond traditional supervised learning by training their models using both labeled and more widely available unlabeled data, such as semi-supervised learning \cite{2li2021dual,4yu2019uncertainty} and weakly-supervised learning \cite{5feng2017discriminative}. However, the performance of these approaches is largely determined by the credibility of the pseudo label. This is because the number of pseudo labels is much larger than ground truth labels. Therefore, the critical question to be answered is how to improve the quality of the pseudo label. To effectively address this issue, we develop a model that can be trained using the medical texts written by domain experts. By learning additional expert knowledge from text information, we can improve the quality of pseudo labels.
\begin{figure*}[]\centering
  \includegraphics[width=0.95\textwidth]{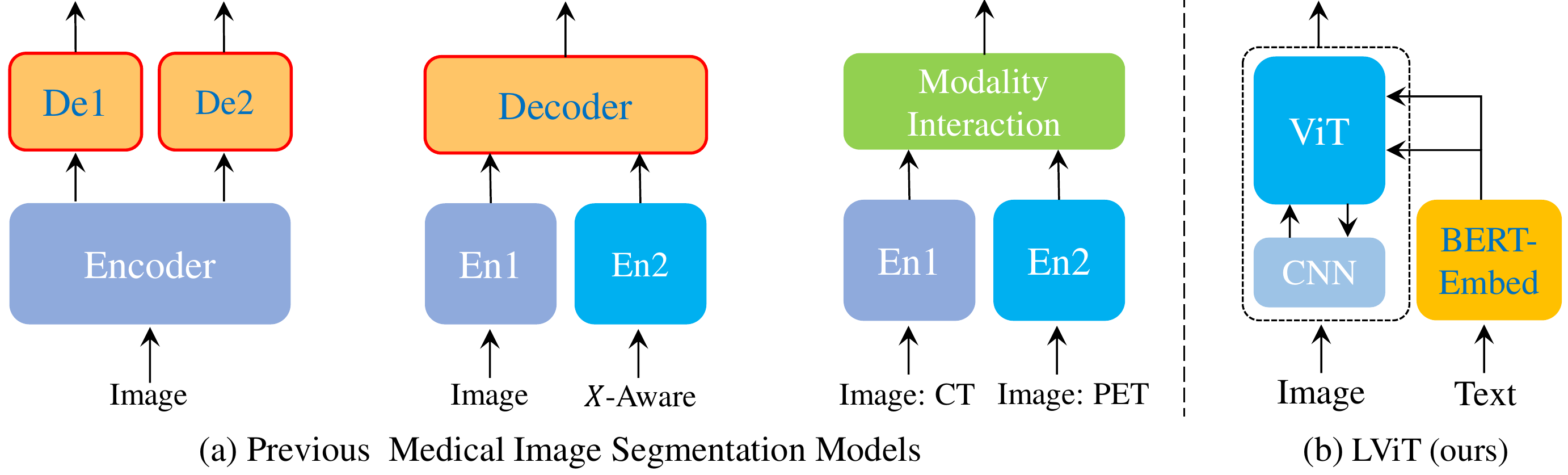}
  \caption{Comparison of current medical image segmentation models and our proposed LViT model.}\label{comp}
  \vspace{-1mm}
\end{figure*}

In summary, \textbf{the challenges exist in two aspects}: \textbf{1)} How to improve the segmentation performance by using the existing image-text information; \textbf{2)} How to make full use of text information to guarantee the quality of pseudo labels. To address \textbf{the first challenge}, 
We propose the LViT model (Figure \ref{comp}(b)), which is innovative in processing images and text. 
In LViT, the text feature vector is obtained by using a embedding layer instead of text encoder, which can  reduce the number of parameters in the model. 
In addition, the hybrid CNN-Transformer structure with Pixel-Level Attention Modules (PLAM) is able to better merge text information and encode global features with Transformer while retaining the CNN's ability to extract local features from images.
To address \textbf{the second challenge}, we design an Exponential Pseudo label Iteration mechanism (EPI) for the proposed LViT, aiming to cross-utilize the label information of labeled data and the latent information of unlabeled data. The EPI indirectly incorporates text information to refine the pseudo label progressively in the way of Exponential Moving Average (EMA) \cite{7grill2020bootstrap}. In addition, the LV (Language-Vision) loss is designed to utilize text information directly to supervise the training of unlabeled medical images. To validate the performance of LViT, we construct three multimodal medical image segmentation datasets containing CT images (MosMedData+ \cite{morozov2020mosmeddata,hofmanninger2020automatic} and ESO-CT) and X-rays (QaTa-COV19 \cite{48degerli2022osegnet}). Results show that LViT has superior segmentation performance, achieving 74.57\% Dice score and 61.33\% mIoU on the MosMedData+ dataset, 83.66\% Dice score and 75.11\% mIoU on the QaTa-COV19 dataset, and 71.53\% Dice score and 59.94\% mIoU on the ESO-CT dataset. And it is worth noting that LViT using 1/4 of the train set labels can still have the same performance as the fully-supervised segmentation method.

\section{Related Work}
\label{gen_inst}
\subsection{Semantic segmentation of medical images}
Semantic segmentation can be considered as the work for pixel-level image classification, and thus many image classification networks have been extended \cite{8long2015fully, 9chen2017deeplab, xu2022mrdff,xu2021multi} to implement semantic segmentation, with fully convolutional network (FCN) \cite{8long2015fully} being commonly considered as the first end-to-end pixel-to-pixel network for semantic segmentation \cite{Yuan2022}. Among them, U-Net \cite{14ronneberger2015u} is considered as a pioneer in medical image segmentation. Based on this, UNet++ \cite{15zhou2018unet++} improved the skip connection of U-Net. However, most of the above methods are very sensitive to the quantity and quality of the data, resulting in limited generalization performance of the models. Therefore, some approaches \cite{2li2021dual,li2021gt,21liu2020semi,li2022semi,20xia2020uncertainty} have explored the application of semi-supervised learning in different areas. The problem of lacking data and its annotation can also be further mitigated by introducing multiple modalities into the learning models. 

\subsection{Vision-language model}
\textcolor{black}{
CLIP \cite{27radford2021learning} is a pioneering work of large-scale vision-language pretrainning (VLP) model, which utilized  contrast learning to learn image representations on a dataset of 400 million pairs (image, text) from scratch.  
By simplifying the processing of visual inputs compared to CLIP, Kim et al. \cite{28kim2021vilt} proposed  a more parameter-efficient architecture, i.e. ViLT, which allows exploiting the power of interaction layers to process visual features while lacking a separate deep visual embedder.
Subsequently, there is a rich line of works on image segmentation \cite{Yang22CVPR,31ding2021vision,Ding22PAMI,32yin2022devil,33xu2022groupvit,li2021gt} that have begun to use text information to improve the segmentation capabilities of models. 
For instance, Ding et al. \cite{Ding22PAMI} developed a Vision-Language Transformer (VLT) framework for referring segmentation by facilitating deep interactions among multi-modal information and fusing linguistic and visual features.  Different from VLT,  Language-Aware Vision Transformer (LAVT) \cite{Yang22CVPR} framework adopted an early fusion scheme for integrating linguistic features into visual features via a pixel-word attention mechanism, which can effectively exploit the Transformer encoder for modeling multi-modal context.
Inspired by VLP in natural image, a few studies have started to utilize text information for assisting with medical image analysis \cite{60bhalodia2021improving, 62muller2021joint,tomar2022tganet}. 
However, compared with natural image, medical image has its own characteristics. 
Unlike the natural image, boundaries between different regions in the medical image are often blurred, and the small gray-scale value differences in the vicinity of the boundaries makes it difficult to extract highly accurate segmentation boundaries. Therefore, it is not applicable to directly apply vision-language model in the natural images to medical image analysis.
VTL and LAVT focus on addressing reference segmentation of the natural images, requiring language encoder for the explicit alignment of image and text. Due to the essential differences between medical images and natural images, it is very difficult to achieve strict alignment between text and image in medical images.
Similarly, ViLT is designed to address multimodal problems, such as Visual Question Answering (VQA) and retrieval tasks. In addition, ViLT is a pure Transformer model without convolution or region supervision for extracting local features, which is not suitable for medical image segmentation with blurred boundaries.
On the other hand, LViT is specifically tailored to medical image segmentation problems. In the LViT model, only the Embedding layer is utilized to transform text features, which requires fewer parameters and lower computational cost. 
In addition, the hybrid CNN-Transformer structure enables us to retain both local and global features of the image, which is crucial for extracting highly accurate segmentation boundaries on medical images. Furthermore, to address the scarcity of medical image labels, an LV (Language-Vision) loss is designed to supervise the training of unlabeled images using text information directly.
}
\subsection{Attention mechanism}
Starting from RAN\cite{34wang2017residual}, researchers have begun to introduce attention mechanisms into the field of computer vision. Woo et al \cite{36woo2018cbam} proposed a well-known Convolutional Block Attention Module (CBAM), where both spatial and channel attentions are used to perform adaptive feature refinement. 
In addition to the original attention mechanisms \cite{huang2020deep}, self-attention mechanisms \cite{30vaswani2017attention,42dosovitskiy2020image,li2021agmb,shan2023coarse} have also begun to enter the field of computer vision. However, since self-attention was originally proposed for solving NLP problems, it faces many challenges, such as high computational cost and neglecting local features of images. Therefore, we propose PLAM to compensate for the lack of attention to local features through self-attention. It also helps the convolutional layer to produce a more effective representation of local features. 
\textcolor{black}{And to address the high computational problem, we utilize the uniform encoder to encode the vision and language features instead of using separate encoders. }

\begin{figure*}[]\centering
  \includegraphics[width=0.95\textwidth]{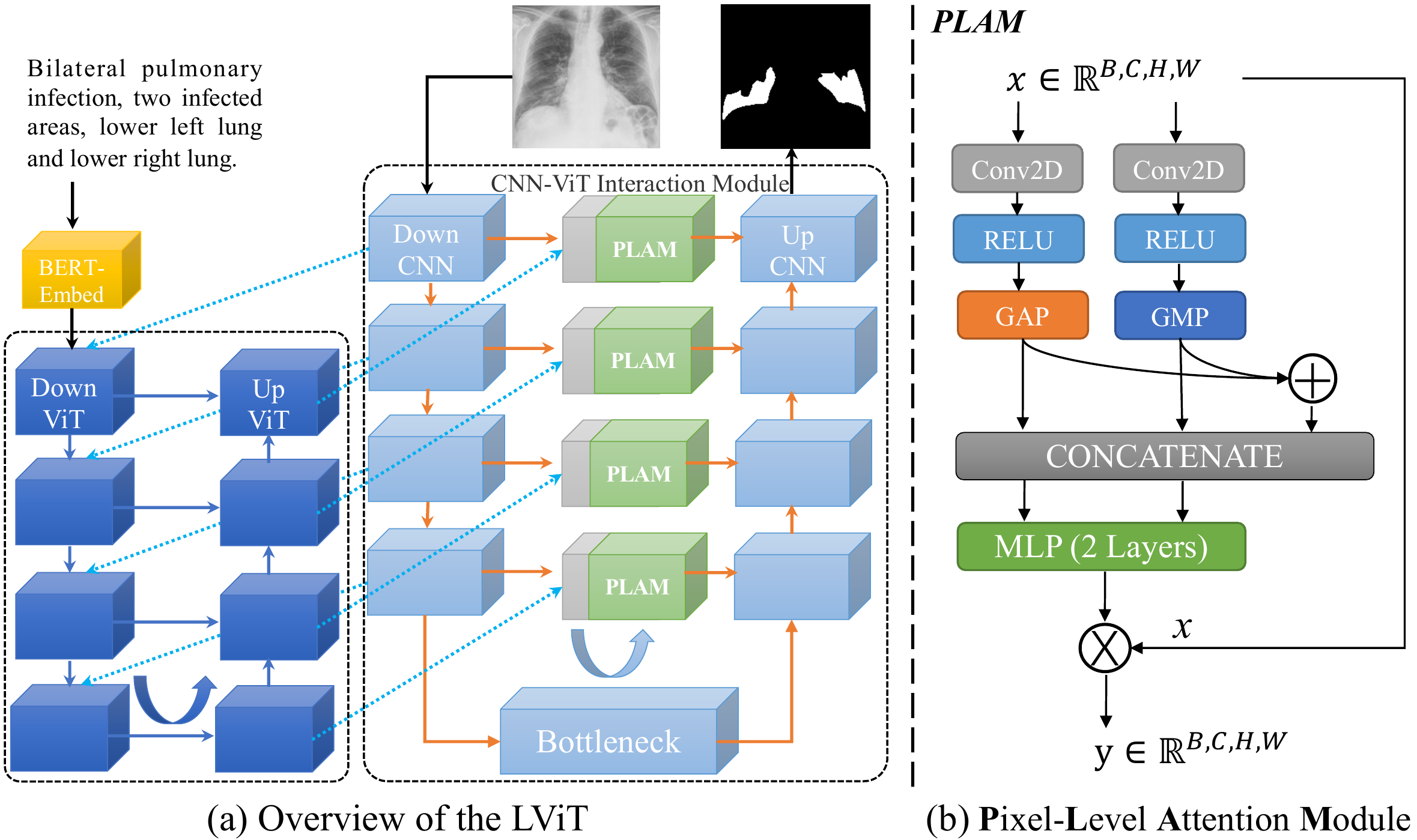}
  \caption{Illustration of (a) the proposed LViT model, and (b) the Pixel-Level Attention Module (PLAM). The proposed LViT model is a Double-U structure formed by combining a U-shape CNN branch with a U-shaped ViT branch.}
  \label{LViT}
  \vspace{-1mm}
\end{figure*}





\section{Method}
\label{others}
As shown in Figure \ref{LViT}, the proposed LViT model is a Double-U structure consisting of a U-shaped CNN branch and a U-shaped Transformer branch.
The CNN branch acts as the source of information input and the segmentation head of prediction output, and the ViT branch is used to merge image and text information, where we exploit the ability of Transformer to process cross-modality information. After a simple vectorization of the text, the text vector is merged with the image vector and send to the U-shaped ViT branch for processing. \textcolor{black}{In the model inference stage, we also need to perform similar processing on text input.} And we pass the fusion information of corresponding size back to the U-shape CNN branch at each layer for the final segmentation prediction. In addition, a Pixel-Level Attention Module (PLAM) is set at the skip connection position of the U-shape CNN branch. With PLAM, LViT is able to retain as much local feature information of images as possible. We also conduct ablation experiments to demonstrate the effectiveness of each module.

\subsection{LViT Model}
\subsubsection{U-shape CNN Branch}
As shown in Figure \ref{LViT}(a), the U-shaped CNN branch is used to receive the image information and act as segmentation head to output the prediction mask. The Conv, BatchNorm(BN), and ReLU activation layers are utilized to compose each CNN module. And image feature are downsampled between each DownCNN module using the MaxPool layer. Concatenate operation is performed between each UpCNN module. The specific process of each CNN module is described by Eqn. \ref{eq3-1} and \ref{eq3-2},
\begin{eqnarray}
    &\operatorname{D}_{i}=\operatorname{DownCNN}_{i}=\operatorname{Relu}\left(B N_{i}\left(\operatorname{Conv}_{i}(\cdot)\right)\right)\vspace{1ex}
    \label{eq3-1}
    \\
    &Y_{\text {DownCNN}, i+1}=\operatorname{MaxPool}\left(\operatorname{D}_{i}\left(Y_{\text{DownCNN},i}\right)\right)
    \label{eq3-2}
\end{eqnarray}
where $Y_{\text{DownCNN},i}$ represents the input of the $i$-th DownCNN module, which becomes $Y_{\text{DownCNN},i+1}$ after the downsampling of the $i$-th DownCNN module and the MaxPool layer. In addition, we design the CNN-ViT interaction module using methods such as upsampling to align the features from ViT, as the details of CNN-ViT interaction module are shown in the appendix. The reconstructed ViT features are also connected with CNN feature by residuals to form CNN-ViT interaction features. In addition, to further improve the segmentation capability for local features, PLAM is designed at the skip connection in the U-shaped CNN branch. So the CNN-ViT interaction features will be fed into PLAM, then the interaction features are transferred to the UpCNN module to give the upward information layer by layer.

\subsubsection{U-shape ViT Branch}
Referring to the U-shaped CNN branch, the U-shaped ViT branch is designed for merging image features and text features. As shown in Figure \ref{LViT}(a), the first layer DownViT module receives the text feature input from BERT-Embed \cite{56devlin2018bert} and the image feature input from the first layer DownCNN module. The pretraining model of the BERT-Embed is the BERT\_12\_768\_12 model, which can convert a single word into a 768-dimensional word vector. The specific cross-modal feature merging operation is expressed by the following equations,
\vspace{-2mm}
\begin{eqnarray}
    \setlength{\belowdisplayskip}{3pt}    
    &Y_{\text {DownViT}, 1}=\operatorname{ViT}\left(x_{\text {img, } 1}+\operatorname{CTBN}\left(x_{\text {text}}\right)\right)
    \label{eq1}
    \\
    &x_{\text {img, } i}=\text {PatchEmbedding}\left(Y_{\text {DownCNN}, i}\right)
    \label{eq2}
    \\
    &x^{\prime}=\operatorname{ViT}_{1}(x)=\operatorname{MHSA}(\operatorname{LN}(x))+x
    \label{eq3}
    \\
    &Y=\operatorname{ViT}_{2}\left(x^{\prime}\right)=\operatorname{MLP}\left(\operatorname{LN}\left(x^{\prime}\right)\right)+x^{\prime}
    \label{eq4}
    \setlength{\abovedisplayskip}{3pt}
\end{eqnarray}
where $x_{img,i}$ represents the image features from DownCNN, $x_{text}$ represents the text features, and PatchEmbedding can help $Y_{DownCNN,i}$ form the embedding features $x_{img,i}$. ViT represents the Transformer encoder\cite{42dosovitskiy2020image}, i.e., $Y=V i T(x)=V i T_{2}\left(V i T_{1}(x)\right)$. ViT consists of the Multi-headed Self-attention (MHSA) module and the MLP layer. And LN represents the normalization layer. The CTBN block also consists of the Conv layer, BatchNorm layer, and ReLU activation layer for aligning the feature dimensions of $x_{img,1}$ and $x_{text}$. The subsequent DownViT modules of layers 2, 3, and 4 receive both feature from the upper DownViT module and the feature from the DownCNN module of the corresponding layer, as shown in Eqn. \ref{eq5}, 
\begin{equation}
    Y_{\text {DownViT }, i+1}=\operatorname{ViT}\left(Y_{\text {DownViT }, i}+x_{i m g, i+1}\right)
    \label{eq5}
\end{equation}
where i=1,2,3. The features of the corresponding size are then transferred back to the CNN-ViT interaction module through the UpViT module. And the feature is merged with the feature from the DownCNN module of the corresponding layer. This will maximize the extraction of image global features and avoid the oscillation of the model performance due to the inaccuracy of text annotation.
\begin{figure}[ht]\centering
  \includegraphics[width=\columnwidth]{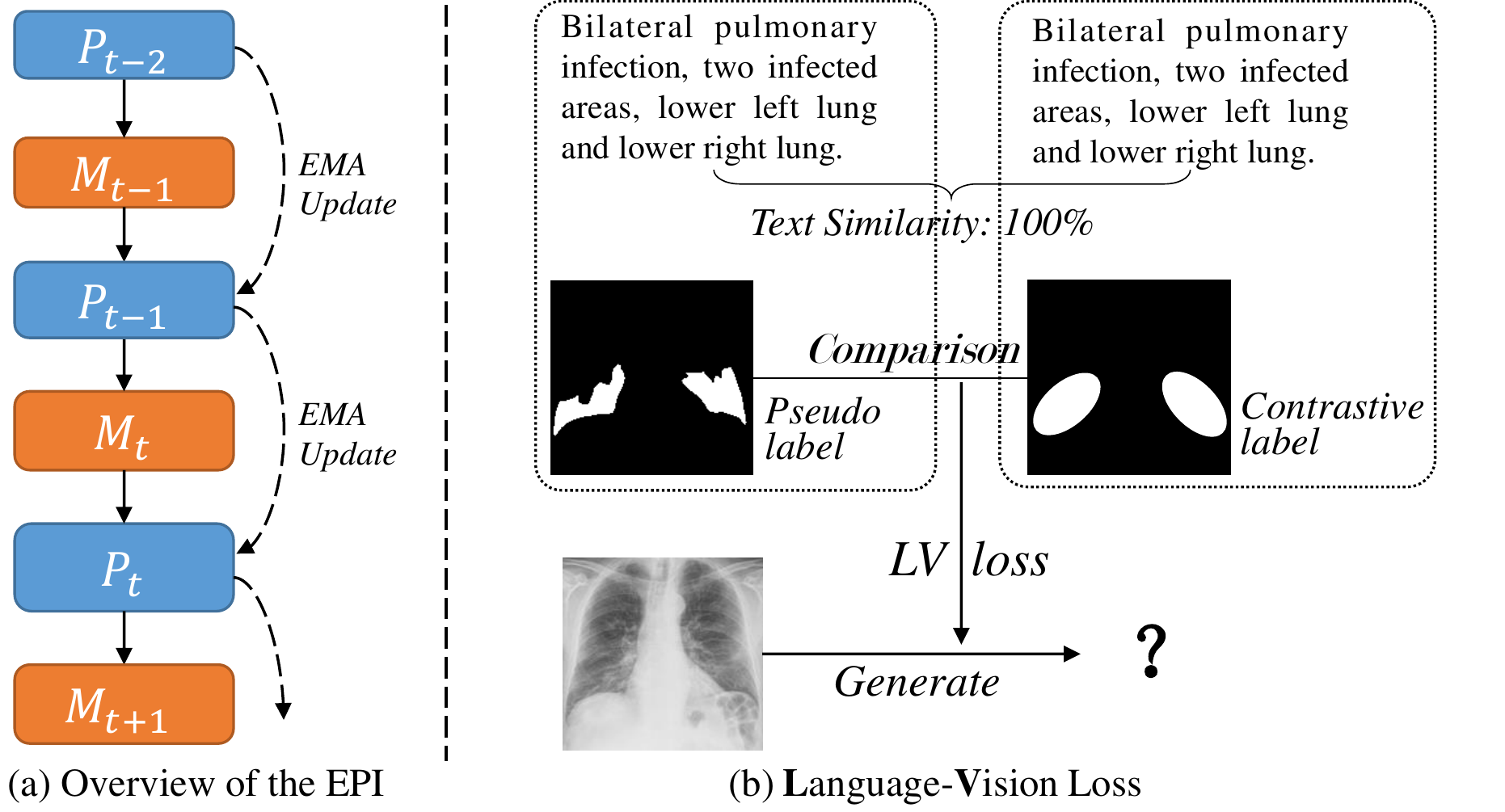}
  \caption{Illustration of (a) Exponential Pseudo-label Iteration mechanism (EPI), and (b) LV (Language-Vision) Loss.}
  \label{EPI&LV}
  \vspace{-3mm}
\end{figure}
\subsubsection{\textbf{P}ixel-\textbf{L}evel \textbf{A}ttention \textbf{M}odule (PLAM)}

As shown in Figure \ref{LViT}(b), \textcolor{black}{PLAM is designed to preserve the local features of images and further merge the semantic features in the text. Besides, it can enhance the performance of the convolutional layer in generating a powerful representation of local features.} Referring to CBAM\cite{36woo2018cbam}, our PLAM uses parallel branches for Global Average Pooling (GAP) and Global Max Pooling (GMP). We also incorporate the concatenate and add operations. The add operation will help merge the corresponding channel features with similar semantics and save computation. In contrast, the concatenate operation can integrate the feature information more intuitively and help preserve the original features of each part. After concatenating the feature information, we use the MLP structure and the multiplication operation to help align the feature size.
\textcolor{black}{
Generally, our PLAM differs in several aspects from the Pixel-word attention module (PWAM) in LAVT \cite{Yang22CVPR}. Firstly, PLAM is designed to enhance local features to mitigate the preference for global features brought by Transformer. In contrast, PWAM is designed to align visual and linguistic representation using cross-attention. Secondly, in terms of implementation,  PLAM utilizes a combination of channel attention and spatial attention, while PWAM uses cross self-attention mechanism. Overall, PLAM aims to enhance local features for the purpose of improving the performance of medical image segmentation with text information. On the other hand, PWAM is designed to align the multimodal features for achieving better referring segmentation results. 
}

\subsection{\textbf{E}xponential \textbf{P}seudo-label \textbf{I}teration mechanism}
In this section, we propose the \textbf{E}xponential \textbf{P}seudo label \textbf{I}teration mechanism (EPI), which is designed to help extend the semi-supervised version of LViT. In EPI, the pseudo label is iteratively updated using the idea of EMA \cite{7grill2020bootstrap}, as shown in Figure \ref{EPI&LV}(a) and Eqn. \ref{3-9},
\begin{equation}
    \setlength{\belowdisplayskip}{3pt}
    P_{t}=\beta \cdot P_{t-1}+(1-\beta) \cdot P_{t}
    \setlength{\abovedisplayskip}{3pt}
    \label{3-9}
\end{equation}
where $P_{t-1}$ represents the prediction of model $M_{t-1}$, and $\beta$ is set as the momentum parameter to 0.99. It is worth noting that here $P_{t-1}$ is an $N$-dimensional prediction vector, where $N$ represents the number of category classes, and each dimension represents the prediction probability. Therefore, EPI can gradually optimize the segmentation prediction results of the model for each unlabeled pixel and be robustness to noisy labels. This is because we do not simply use the pseudo label predicted by one generation of models as the target for the next-generation model, which can avoid sharp deterioration of the pseudo label quality. The theoretical proof for the effectiveness of EPI algorithm is as follows.

\textbf{Proof:} The basic assumption for the EPI algorithm is the model weights will dither around the actual optimum in the last $n$ generations, and therefore the pseudo label predicted by the model will also dither around the mask in the last $n$ generations. We expand $P_{t}$ around $t$ in Eqn. \ref{3-9} to Eqn.\ref{3-10},
\vspace{1mm}
\begin{equation}
    \setlength{\belowdisplayskip}{1pt}
    P_{t}=\beta^{n}P_{t-n}+(1-\beta) \cdot\left(\beta^{n-1} P_{t-n+1}+\cdots+P_{t}\right).
    \setlength{\abovedisplayskip}{1pt}
    \label{3-10}
\end{equation}

In particular, we let $n=1/(1-\beta)$ and $\beta^{n}=\beta^{\frac{1}{1-\beta}} \approx \frac{1}{e}$. So for the first $1/(1-\beta)$ generations, $P_{t}$ decays to a weighted average of $1/e$. Further, we introduce an adjustment gradient $g_{t-1}$ for the predicted label $P_{t-1}$, which leads to Eqn. \ref{3-11},
\begin{equation}
    \setlength{\belowdisplayskip}{1pt}
    P_{t}=P_{t-1}-g_{t-1}=\cdots=P_{1}-\sum_{i=1}^{n-1} g_{i}.
    \setlength{\abovedisplayskip}{1pt}
    \label{3-11}
\end{equation}
Similarly, we extend Eqn. \ref{3-10} when $t=n$ and $P_{0}\approx P_{1}$. 
Comparing with Eqn.\ref{3-11}, it can be seen from Eqn. (\ref{eq11-0} - \ref{eq11-4}) that the EPI algorithm adds the weight coefficient $1-\beta^{n-i}$ for the gradient descent step of the $i$-th iteration. $1-\beta^{n-i}$ will decreases as $i$ increases, so the change of pseudo label is finally stabilized and obtain the pseudo label with high confidence. 

\vspace{-3mm}
\begin{eqnarray}
&&\!\!\!\!\!\!\!\!\!\!\!\!\!\!\!\!\overset{\sim}{P}=\beta^{n} P_{0}\label{eq11-0}\\
&&\!\!\!\!\!\!\!\!\!\!\!\!\!\!\!\! P_{t}=\overset{\sim}{P}+(1-\beta) \cdot\left(\beta^{n-1} P_{1}+\beta^{n-2} P_{2}+\cdots+ P_{n}\right)  \qquad \quad  
    \label{eq11-1}\\
&&\!\!\!\!\!\!\!\!\!\!\!\!\!\!\!\! P_{t}=\overset{\sim}{P}+(1-\beta) \cdot\left(\beta^{n-1} P_{1}+\cdots+\left(P_{1}-\sum_{i=1}^{n-1} g_{i}\right)\right) 
    \label{eq11-2}\\
&&\!\!\!\!\!\!\!\!\!\!\!\!\!\!\!\! P_{t}=\overset{\sim}{P}+(1-\beta) \cdot\left(\frac{1-\beta^{n}}{1-\beta} P_{1}-\sum_{i=1}^{n-1} \frac{\left(1-\beta^{n-i}\right) g_{i}}{1-\beta}\right) 
    \label{eq11-3}\\
&&\!\!\!\!\!\!\!\!\!\!\!\!\!\!\!\! P_{t} \approx P_{1}-\sum_{i=1}^{n-1}\left(1-\beta^{n-i}\right) g_{i}. \label{eq11-4}
\end{eqnarray}

\subsection{LV (Language-Vision) Loss}
To further utilize the text information to guide the pseudo-label generation, we design the LV (Language-Vision) loss function, as shown in Figure \ref{EPI&LV}(b). 
\textcolor{black}{
Generally, the positions of human organs in medical images are relatively fixed. Thus,} we can use structured text information to form the corresponding mask (the contrastive label). And we calculate the cosine similarity between the texts, as shown in Eqn. \ref{3-13},
\begin{equation}
    \text {TextSim}=\frac{x_{\text {text }, p} \cdot x_{\text {text}, c}}{\left|x_{\text {text }, p}\right| \times\left|x_{\text {text }, c}\right|}
    \label{3-13}
\end{equation}
where $x_{text,p}$ represents the text feature vector corresponding to the pseudo label, and $x_{text,c}$ represents the text feature vector corresponding to the contrastive label. After that, according to $TextSim$, we select the contrastive text with the highest similarity and find the segmentation mask corresponding to that text; we calculate the cosine similarity between the predicted segmentation pseudo-label and the contrastive label using the label similarity, as shown in Eqn. \ref{3-14} and \ref{3-15},
\begin{eqnarray}
    &\operatorname{ImgSim}=\frac{x_{img, p} \cdot x_{img,c}}{\left|x_{img,p}\right| \times\left|x_{img,c}\right|}
    \label{3-14}
    \\
    &L_{LV}=1-\operatorname{ImgSim}
    \label{3-15}
\end{eqnarray}
where $x_{img,p}$ represents the pseudo-label feature vector, and $x_{img,c}$ represents the comparison label feature vector. Compared to Euclidean distance, cosine similarity is not sensitive to absolute values and reflects the degree of similarity more qualitatively, consistent with our task motivation. The contrastive labels mainly provide labeling information of the approximate location instead of refinement for the boundaries. Therefore, the primary purpose of LV loss is to avoid mis-segmentation or mislabelled cases with significant differences. For this reason, we only use LV loss in the unlabeled case because the contrastive labels are of little help for performance improvement when the data is labeled. And in case of no label, LV loss with consistency supervision can avoid the sharp deterioration of the pseudo-label quality.
\textcolor{black}{
It is important to note that the Pseudo and Contrastive labels in our LViT aim to address different issues compared to the masked conservative learning in VLT \cite{Ding22PAMI}. Firstly, the Pseudo and Contrastive labels are designed for semi-supervised learning, whereas masked conservative learning aims to explore the knowledge of different language expressions pertaining to a single object. Secondly, LViT determines whether a case is similar by calculating text similarity, while VLT achieves this by extracting text features. However, it is difficult to determine the similarity between radiology reports through implicit feature extraction in the medical field, as different radiology reports may have only a few wording changes. Therefore, structured formats are typically used to differentiate between reports. In addition, different from the masked conservative learning, we design an Exponential Pseudo label Iteration mechanism (EPI) to guarantee the quality of pseudo labels with text information, which utilizes the label information of labeled data and the latent information of unlabeled data in a cross-utilized manner.}

\subsection{Proof of CNN-Transformer structure superiority}
Unlike the previous Vision-and-Language work, our proposed LViT model is innovative in processing images and text. We do not use text encoder and creatively use the interaction between CNN and ViT to extract features.

\textcolor{black}{\textbf{Proof:} For the sake of description, we assume that the patch size in ViT is equal to the kernel size in CNN, which is $S$. The input matrix is $M$, and output of convolution is $Y_{cnn}$.
\vspace{-2mm}
\begin{eqnarray}
    \mathrm{Y}_{cnn,k}(i, j)=\sum_{\xi=0}^{S} \sum_{\eta=0}^{S} f_{k}(\xi, \eta) * M(i-\xi, j-\eta)\\
   {Y}_{c n n}(i, j)=\left[\mathrm{Y}_{c n n, 1}(i, j) ;  \ldots ; \mathrm{Y}_{c n n, C}(i, j)\right]
   \setlength{\belowdisplayskip}{0pt}
\end{eqnarray}
where $f_{k}$ represents the convolution kernel of the $k$-th channel, and $Y_{cnn,k} (i,j)$ represents the output of the $k$-th channel after convolution. The total convolution outputs of $C$ channels form $Y_{cnn}(i,j)$. And the convolution operation $f(x)$ is satisfying shift invariance and scale invariance, i.e., if $Y(x)=f(x)*M(x)$, then we have $Y(x-\delta)=f(x)*M(x-\delta)$, and if $Y(x)=f(x)*M(x)$, then we have $|\delta|Y(x/\delta)=f(x/\delta)*M(x/\delta)$. Therefore, CNNs are good at learning shallow features and are affine-invariant. The kernel size is fixed, so each kernel can only learn one aspect of local information, like points, lines, and boundaries. And since the convolutional kernel $f_{k}(\xi, \eta)$ of each channel shares the weights on the whole image, convolving the whole image with convolutional kernels that focus on boundary features is equivalent to doing whole-image filtering on the image. Similarly, we set the output after multi-head self-attention as $Y_{vit}$, as shown in the Eqn. \ref{eqA-4} and \ref{eqA-5},
   \vspace{-1mm}
    \begin{eqnarray}
    \mathrm{Y}_{v i t, h}=\operatorname{Softmax}\left(\frac{Q_{h}^{T} \cdot \mathrm{K}_{h}}{\sqrt{d}}\right) \cdot \mathrm{V}_{h}  
    \label{eqA-4}\\\vspace{-1mm}
    \mathrm{Y}_{v i t}=L N\left(\left[\mathrm{Y}_{v i t, 1} ; \mathrm{Y}_{v i t, 2} ; \ldots ; \mathrm{Y}_{v i t, H}\right]\right)
       \setlength{\belowdisplayskip}{2pt}
    \label{eqA-5}
    \end{eqnarray}
where ${Y}_{v i t, h}$ denotes the output of the $h$-th self-attention head, and $d$ prevent the feature gradient from vanishing after Softmax. $LN$ represents the linear layer, which aims to reduce the dimensionality of the output features. And $Q_{h}$, $K_{h}$, and $V_{h}$ in the self-attention mechanism are transformations for their own inputs, $\operatorname{Softmax}\left(\frac{Q_{h}^{T} \cdot \mathrm{K}_{h}}{\sqrt{d}}\right) \cdot \mathrm{V}_{h}$ is computing the similarity between them. So self-attention is essentially focusing on the invariance of input features. The input features $M$ are the whole image for ViT, so ViT is easier to learn the global features with more robustness than CNN.}

\begin{table}[!ht]
\caption{The specific division of different datasets.}
\label{table_dataset}
\centering
\resizebox{0.95\columnwidth}{!}{%
\begin{tabular}{c|c|c|c}
\hline
          & \textbf{QaTa-COV19} & \textbf{MosMedData+ } & \textbf{ESO-CT}\\ \hline
Train set & 5716                & 2183 &  182            \\ 
Val set   & 1429                & 273   &  46           \\ 
Test set  & 2113                & 273   & 58           \\ \hline
Total  & 9258                & 2729   & 286           \\ \hline
\end{tabular}
}
\end{table}

\begin{table*}[]
\setlength{\belowcaptionskip}{0.2cm}
\caption{Performance comparison between our method (LViT) and other state-of-the-art methods on the QaTa-COV19 and MosMedData+ datasets. The "W" in LViT-TW refers to without the text information. The "Hybrid" means CNN-Transformer structure.}
\label{table4-1}
\centering
\resizebox{\textwidth}{!}{%
\begin{tabular}{cccccccccc}
\midrule
\multicolumn{1}{c}{} &
  \multicolumn{1}{c}{} &
   &&&&
  \multicolumn{2}{c}{\textbf{QaTa-COV19}} &
  \multicolumn{2}{c}{\textbf{MosMedData+}} \\ \cmidrule(r){7-8} \cmidrule(l){9-10}
\multicolumn{1}{c}{\multirow{-2}{*}{\textbf{Method}}} &\multirow{-2}{*}{\textbf{Backbone}}&\multirow{-2}{*}{\textbf{Text}}&\multirow{-2}{*}{\textbf{Label ratio}}&
  \multirow{-2}{*}{\textbf{Param (M)}} &
  \multirow{-2}{*}{\textbf{Flops (G)}} &
  Dice (\%) &  mIoU (\%) &  Dice (\%) &  mIoU (\%) \\ \hline\rowcolor{black!10}
U-Net \cite{14ronneberger2015u}& CNN & $\times$ & \textcolor{black}{100\%} & 14.8 &  50.3 &  79.02 &  69.46 &  64.60 &  50.73 \\\rowcolor{black!10}
UNet++ \cite{15zhou2018unet++}& CNN & $\times$ & \textcolor{black}{100\%}& 74.5 &  94.6 &  79.62 &  70.25 &  71.75 &  58.39 \\\rowcolor{black!10}
AttUNet \cite{51oktay2018attention}& CNN & $\times$ & \textcolor{black}{100\%}& 34.9 &  101.9 &  79.31 &  70.04 &  66.34 &  52.82 \\\rowcolor{black!10}
nnUNet \cite{52isensee2021nnu}& CNN & $\times$ & \textcolor{black}{100\%}& 19.1 &  412.7 &  80.42 &  70.81 &  \textbf{72.59} &  60.36 \\\rowcolor{black!10}
TransUNet \cite{54chen2021transunet}& Hybrid & $\times$ & \textcolor{black}{100\%}& 105 &  56.7 &  78.63 &  69.13 &  71.24 &  58.44 \\\rowcolor{black!10}
Swin-Unet \cite{55cao2021swin}& Hybrid & $\times$ & \textcolor{black}{100\%}& 82.3 &  67.3 &  {78.07} &  {68.34} &  63.29 &  50.19 \\\rowcolor{black!10}
UCTransNet \cite{50wang2021uctransnet}& Hybrid & $\times$ & \textcolor{black}{100\%}& 65.6 &  63.2 &  79.15 &  69.60 &  65.90 &  52.69 \\\hline \rowcolor{black!15}
\textbf{LViT-TW (1/4)}& Hybrid & $\times$  & \textcolor{black}{25\%}& 28.0 &  54.0 &  79.08 &  69.42 & 70.65  & 58.07  \\ \rowcolor{black!15}
\textbf{LViT-TW (1/2)}& Hybrid & $\times$  & \textcolor{black}{50\%} & 28.0 &  54.0 &  {80.35} &  {70.74} & 71.89 & 59.63  \\ \rowcolor{black!15}
\textbf{LViT-TW}& Hybrid & $\times$  & \textcolor{black}{100\%} & 28.0 &  54.0 &  \textbf{81.12} &  \textbf{71.37} &  72.58 &
  \textbf{60.40} \\ 
\hline
\rowcolor{yellow!10}
ConVIRT\cite{zhang2020contrastive}& CNN &\checkmark& \textcolor{black}{100\%} & 35.2 & 44.6  &79.72 &70.58 &72.06 &59.73\\ \rowcolor{yellow!10}
TGANet\cite{tomar2022tganet}& CNN & \checkmark & \textcolor{black}{100\%} & 19.8 & 41.9 & 79.87 & 70.75 & 71.81 & 59.28 \\\rowcolor{yellow!10}
CLIP\cite{27radford2021learning} & Hybrid & \checkmark & \textcolor{black}{100\%} & 87.0& 105.3& 79.81& 70.66& 71.97 &59.64\\\rowcolor{yellow!10}
GLoRIA\cite{huang2021gloria} & Hybrid & \checkmark & \textcolor{black}{100\%} &45.6 & 60.8&79.94 &70.68 & 72.42&60.18 \\\rowcolor{yellow!10}
\textcolor{black}{ViLT\cite{28kim2021vilt}} & \textcolor{black}{Hybrid} & \textcolor{black}{\checkmark} & \textcolor{black}{100\%} &\textcolor{black}{87.4}& \textcolor{black}{55.9}&\textcolor{black}{79.63}&\textcolor{black}{70.12}&\textcolor{black}{72.36}&\textcolor{black}{60.15}
\\\rowcolor{yellow!10}
\textcolor{black}{LAVT\cite{Yang22CVPR}} & \textcolor{black}{Hybrid} & \textcolor{black}{\checkmark} & \textcolor{black}{100\%} &\textcolor{black}{118.6}& \textcolor{black}{83.8}&\textcolor{black}{79.28} & \textcolor{black}{69.89}& \textcolor{black}{73.29}&\textcolor{black}{60.41}
\\\hline \rowcolor{yellow!15}
\textbf{LViT-T (1/4)}& Hybrid & \checkmark & \textcolor{black}{25\%} & 29.7 &  54.1 &  {80.95} &  {71.31} & 72.48 & 60.31 \\  \rowcolor{yellow!15}
\textbf{LViT-T (1/2)}& Hybrid & \checkmark & \textcolor{black}{50\%} &  29.7 &  54.1 &  {82.73} &  {73.99} &  73.56 & 61.05 
   \\ \rowcolor{yellow!15}
\textbf{LViT-T}& Hybrid & \checkmark & \textcolor{black}{100\%} &  29.7 &  54.1 & \textbf{83.66} &  \textbf{75.11} &  \textbf{74.57} &  \textbf{61.33} \\ 
\hline
\end{tabular}%
}
\end{table*}

\section{Experiments}
\label{exp}
\subsection{Setup}
Three datasets are used in the experiments to evaluate the performance of our method. The first one is the \renewcommand{\thefootnote}{\arabic{footnote}}{MosMedData+\footnote{http://medicalsegmentation.com/covid19}} dataset \cite{morozov2020mosmeddata,hofmanninger2020automatic}, which contains 2729 CT scan slices of lung infections. The second one is the QaTa-COV19 dataset \cite{48degerli2022osegnet}, which is compiled by researchers from Qatar University and Tampere University. This dataset consists of 9258 COVID-19 chest X-ray radiographs with manual annotations of COVID-19 lesions for the first time. In addition, text annotations for the datasets are extended by us to be used for training the vision-language model. We extend text annotations on the QaTa-COV19 dataset for the first time with the help of professionals. The text annotations focuse on whether both lungs are infected, the number of lesion regions, and the approximate location of the infected areas. 
For example, \textbf{"Bilateral pulmonary infection, two infected areas, upper left lung and upper right lung."} refers to bilateral lung infection, and there are two infection areas located in the upper left lung and the upper right lung respectively. 
The text annotations on MosMedData+ dataset mainly contain the same information as QaTa-COV19 dataset, and the text structure is similar, e.g., \textbf{"Unilateral pulmonary infection, two infected areas, middle lower left lung."}. 
The third dataset is the ESO-CT dataset, which consists of 286 cases, and the detail will be presented in the section of generalization study.
\textcolor{black}{
Those text annotations were provided and verified by two professionals from the Department of Radiation Oncology, UT Southwestern Medical Center. The radiologists independently annotated the same image, and then we compared their annotations to ensure consistency. Additionally, we conducted a quality check based on the provided mask to ensure that there was no excessive deviation in the text annotations.}
The loss function we use is shown in Eqn. \ref{4-20}, where $L_{Dice}$ means dice loss and $L_{CE}$ means cross-entropy loss. For the unlabeled data, an additional term on the loss $L_{LV}$ is introduced with $\alpha=0.1$. And for the labeled data, $\alpha=0$.
Dice and mIoU are used to evaluate the segmentation performance. And early stop mechanism is used during training phase.
\begin{eqnarray}
\setlength{\belowdisplayskip}{3pt}
&L_{Dice}=1-\sum_{i=1}^{N}\sum_{j=1}^{C}{\frac{1}{NC}\cdot\frac{2\left|p_{ij}\cap y_{ij}\right|}{\left(\left|p_{ij}\right|+\left|y_{ij}\right|\right)}}
\label{4-17}
\\
&L_{CE}=-\sum_{i=1}^{N}\sum_{j=1}^{C}{\frac{1}{N}\cdot}y_{ij}\log{\left(p_{ij}\right)}
\label{4-18}
\\
&L_{sup}=(L_{Dice}+L_{CE})/2
\label{4-19}
\\
&L_{unsup}=(L_{Dice}+L_{CE})/2+\alpha\cdot\ L_{LV}\ 
\label{4-20}
\setlength{\abovedisplayskip}{3pt}
\end{eqnarray}
where $N$ represents the number of pixels, $C$ represents the number of categories, which is set to 1 in our experiments. $p_{ij}$ represents the prediction probability that pixel $i$ belongs to category $j$, $y_{ij}$ represents whether pixel $i$ belongs to category $j$. If pixel $i$ belongs to category $j$, then $y_{ij}$ is 1, otherwise 0.

\subsection{Evaluation Metrics}
For the evaluation metrics, the Dice score and the mIoU metric are used to evaluate the performance of our LViT model and other SOTA methods, as shown in Eqn. \ref{4-21} and \ref{4-22},
\begin{eqnarray}
&\setlength{\belowdisplayskip}{3pt}
Dice=\sum_{i=1}^{N}\sum_{j=1}^{C}{\frac{1}{NC}\cdot\frac{2\left|p_{ij}\cap y_{ij}\right|}{\left(\left|p_{ij}\right|+\left|y_{ij}\right|\right)}}=1-L_{Dice}
\label{4-21}
\\
&mIoU=\sum_{i=1}^{N}\sum_{j=1}^{C}{\frac{1}{NC}\cdot\frac{|p_{ij}\cap y_{ij}|}{|p_{ij}\cup y_{ij}|}}
\label{4-22}
\setlength{\abovedisplayskip}{3pt}
\end{eqnarray}
where $N$ represents the number of pixels, $C$ represents the number of categories, $p_{ij}$ and $y_{ij}$ also have the same definition as in the above section.
\vspace{-2mm}
\subsection{Implementation Details}
Our proposed approach is implemented using Pytorch. The main parameters of the server are listed below: the operating system is Ubuntu 16.04.12 LTS, the CPU is Intel(R) Xeon(R) Gold 5218, the GPU is a 2-card TESLA V100 32G, and the memory capacity is 128 GB. 
In terms of dataset division, we split the train set and the validation set from the original train set. Then, the train set is divided into labeled and unlabeled train sets in a specific ratio. The number of samples in each dataset is presented in Table \ref{table_dataset}.

\begin{figure*}[!ht]
\centering
  \includegraphics[width=\textwidth]{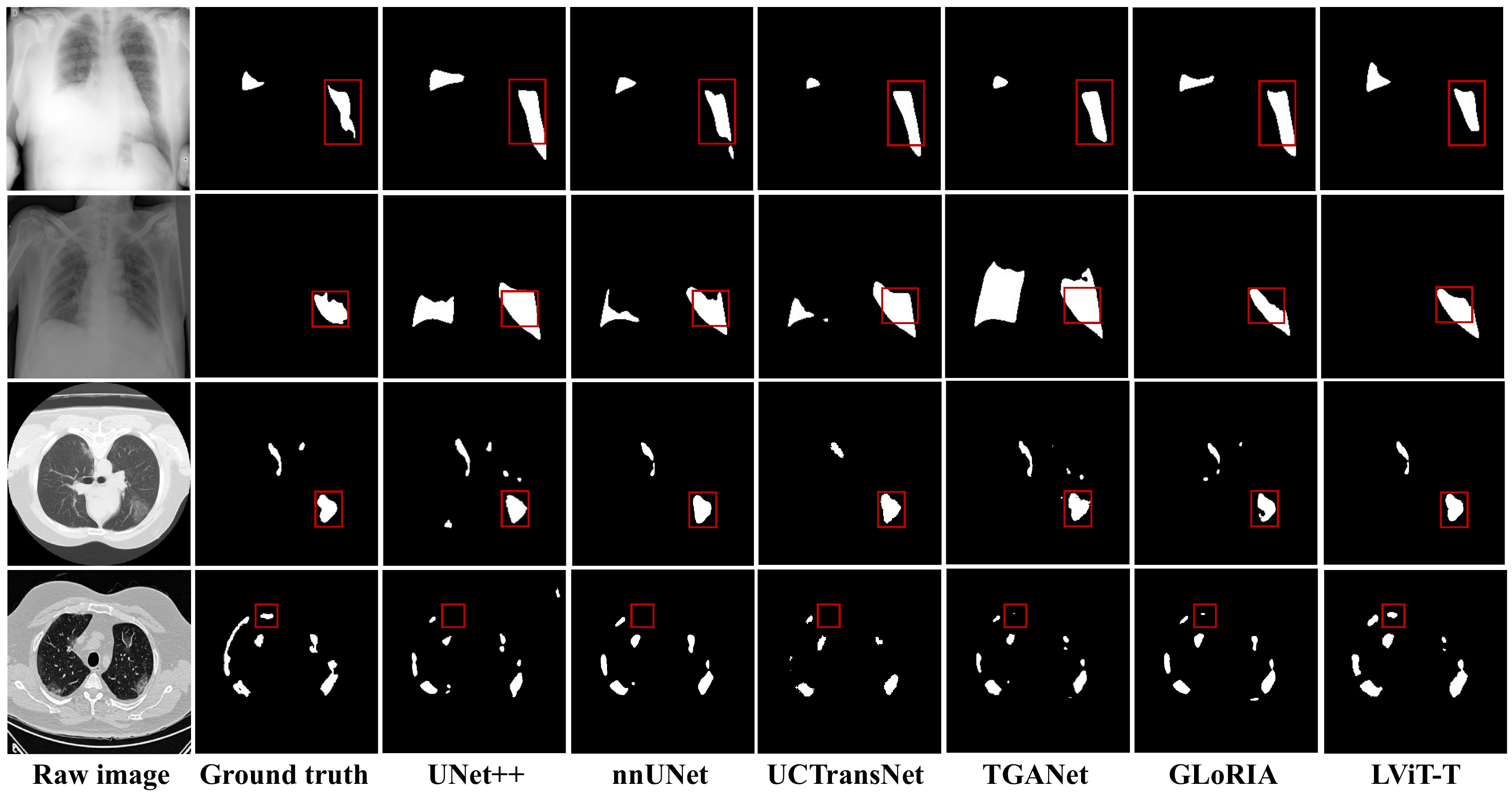}
  \caption{\textcolor{black}{Qualitative results on the QaTa-COV19 and the MosMedData+ datasets.}}
  \label{ExpToSOTA}
  \vspace{-3mm}
\end{figure*}
The initial learning rate is set to 3e-4 for the QaTa-COV19 dataset and 1e-3 for the MosMedData+ dataset. We also use an early stop mechanism until the performance of model does not improve for 50 epochs. Different batch sizes are also set for each dataset since they have different data size. The default batch size is 24 for the QaTa-COV19 dataset and the MosMedData+ dataset.

\begin{table*}[!ht]
\caption{Ablation study on the effectiveness of supervised components: DownViT, UpViT, PLAM, Text \& semi-supervised components: EPI, Text, Loss $L_{LV}$ on the QaTa-COV19 dataset.}
\label{table4-2}
\centering
\resizebox{\textwidth}{!}{
\begin{tabular}{ccccccccccc}
\toprule
\textbf{Method} & \textbf{CNN} & \textbf{DownViT} & \textbf{UpViT} & \textbf{\textcolor{black}{GAP/PLAM}} & \textbf{\textcolor{black}{GMP/PLAM}} & \textbf{Text} & \textbf{EPI}& \bm{${\operatorname{L}_{LV}}$}& Dice (\%) & mIoU (\%) \\ \midrule
nnUNet                 & \checkmark  &   &   &   &   & & & & 80.42         & 70.81         \\
\midrule
\multirow{7}{*}{LViT-T} & \checkmark & \checkmark &   &   &   & & & &80.73          & 70.96          \\
                      & \checkmark & \checkmark & \checkmark &  & & & &  & 80.85          & 71.12          \\
                      & \textcolor{black}\checkmark & \textcolor{black}\checkmark & \textcolor{black}\checkmark & \textcolor{black}\checkmark &  &  & & & \textcolor{black}{81.03} & \textcolor{black}{71.29}          \\
                      & \textcolor{black}\checkmark & \textcolor{black}\checkmark & \textcolor{black}\checkmark &  & \textcolor{black}\checkmark &  & & & \textcolor{black}{80.92} & \textcolor{black}{71.21}          \\
                      & \checkmark & \checkmark & \checkmark & \checkmark & \checkmark &  & & & 81.12          & 71.37          \\
                      &   & \checkmark & \checkmark & \checkmark & \checkmark & \checkmark & & & 80.52          & 70.43          \\
                      & \checkmark & \checkmark & \checkmark & \checkmark & \checkmark & \checkmark & & & \textbf{83.66} & \textbf{75.11} \\  
\midrule
\multirow{4}{*}{LViT-T (1/4)} & \textcolor{black}\checkmark & \textcolor{black}\checkmark & \textcolor{black}\checkmark & \textcolor{black}\checkmark & \textcolor{black}\checkmark &  &  &   & \textcolor{black}{78.87} & \textcolor{black}{69.25}
\\& \textcolor{black}\checkmark & \textcolor{black}\checkmark & \textcolor{black}\checkmark & \textcolor{black}\checkmark & \textcolor{black}\checkmark & \textcolor{black}\checkmark &  &   & \textcolor{black}{80.41} & \textcolor{black}{70.79}
\\& \checkmark & \checkmark & \checkmark & \checkmark & \checkmark &  & \checkmark &   & 79.08                         & 69.42 
\\& \checkmark & \checkmark & \checkmark & \checkmark & \checkmark & \checkmark & \checkmark &  &   80.67          & 71.08          
\\& \checkmark & \checkmark & \checkmark & \checkmark & \checkmark
                           & \checkmark &\checkmark &  \checkmark &  \textbf{80.95} & \textbf{71.31}\\\midrule
\textcolor{black}{LViT-T (1/4, sup with $L_{LV}$)} & \textcolor{black}{\checkmark} & \textcolor{black}\checkmark & \textcolor{black}\checkmark & \textcolor{black}\checkmark & \textcolor{black}\checkmark & \textcolor{black}\checkmark & \textcolor{black}\checkmark & \textcolor{black}\checkmark & \textcolor{black}{80.98} & \textcolor{black}{71.30} \\\bottomrule
\end{tabular}
}
\vspace{-2mm}
\end{table*}
\vspace{-3mm}
\subsection{Comparison with State-of-the-Art Methods}
We compare the performance of our LViT model with
several CNN and Transformer based segmentation models. The number of network parameters and the computational cost of different methods are also reported. Note that the numbers after the methods refer to the ratio of labels used, e.g., 
LViT-T (1/2) refers to the experimental results of the LViT-T model using 1/2 of the train set labels. And LViT-T means the \textbf{Tiny} version of LViT. The quantitative experimental results are listed in Table \ref{table4-1}.
Experimental results on the QaTa-COV19 dataset show that LViT-TW/ LViT-T is able to achieve better performance than the previous SOTA method with smaller number of parameters and lower computational cost. In detail, LViT-T improves the Dice score by 3.24\% and the mIoU score by 4.3\% compared to the suboptimal nnUNet. It is also worth noting that LViT-T still outperforms other SOTA methods even when only 1/4 of the training labels are used. Similarly, it can be seen that LViT-T has a 2.54\% higher Dice score and a 4.05\% better mIoU score than LViT-TW. This also indicates that introducing text information is able to improve model performance effectively. A similar trend is observed for the MosMedData+ dataset. On the MosMedData+ dataset, compared to GLoRIA, LViT-T improves the Dice value by 2.15\% and the mIoU value by 1.15\%. Even LViT-TW and LViT-T (1/4) can achieve comparable performance to nnUNet and UCTransNet.

The qualitative results of our model and other state-of-the-art methods on the MosMedData+ and QaTa-COV19 datasets are shown in Figure \ref{ExpToSOTA}, where four baseline methods are selected for comparison.
Qualitative results shows that LViT-T has excellent semantic segmentation capabilities, especially when compared to other SOTA methods where the mis-segmentation phenomenon is greatly reduced. As can be seen from the red boxes in Figure \ref{ExpToSOTA}, UNet++, nnUNet and TransUNet all have more severe mis-segmentation than LViT. It also shows the introduction of text information in our learning mechanism can better guide the training of the model, and consequently lead to more accurate segmentation. In addition, compared with different multimodal segmentation methods, LViT also has obvious advantages as can be shown in Table \ref{table4-1} and Figure \ref{ExpToSOTA}. Thanks to the benefits brought by the integration of text and image information in the same encoder, LViT is more delicate in the segmentation boundary.

\begin{table*}[!ht]
\caption{Ablation study on different Model Sizes: LViT-T, LViT-S, LViT-B. The Dice and IoU are in
‘mean±std’ format. \textcolor{black}{The std stands for Standard Deviation in three times runs.}}
\label{table4-4}
\centering
\resizebox{0.95\textwidth}{!}{%
\tiny
\begin{tabular}{lcccccc}
\toprule
\multirow{2}{*}{\textbf{Model Size}} & \multirow{2}{*}{Param(M)} & \multirow{2}{*}{Flops(G)} & \multicolumn{2}{c}{\textbf{QaTa-COV19}} & \multicolumn{2}{c}{\textbf{MosMedData+}} \\ \cmidrule(r){4-5} \cmidrule(l){6-7}
                                     &                           &                           & Dice (\%)     & mIoU   (\%)    & Dice (\%)    & mIoU (\%)    \\ \cmidrule{1-7}
LViT-TW & 28.0 & 54.0&
  81.12$\pm$1.9 & 71.37$\pm$2.4 & 72.58$\pm$1.4 & 60.40$\pm$0.7  \\
LViT-SW                         &            53.1               &            63.8        & 81.54$\pm$1.7         & 71.91$\pm$2.0   & 72.75$\pm$1.2 & 60.52$\pm$0.6 \\
LViT-BW                         &          69.8        &          70.4          & \textbf{81.59$\pm$1.6}         & \textbf{71.82$\pm$2.1}          & \textbf{72.84$\pm$1.1} & \textbf{60.58$\pm$0.6}  \\ \cmidrule{1-7}
LViT-T                               &         29.7                  &            54.1               & \textbf{83.66$\pm$0.8}         & 75.11$\pm$1.4          &      74.57$\pm$0.8  &      61.33$\pm$0.5   \\
LViT-S                               &         54.8                  &            63.9         & 83.41$\pm$1.0         & 74.84$\pm$1.3          &  74.65$\pm$0.7  &    61.46$\pm$0.5   \\
LViT-B                               &         71.5                  &           70.5          & 83.63$\pm$0.9         & \textbf{75.28$\pm$1.1}  &  \textbf{74.76$\pm$0.5}  &    \textbf{61.53$\pm$0.4}  \\
 \midrule
 \vspace{-4mm}
\end{tabular}
}
\end{table*}

\begin{table*}[!ht]
\caption{\textcolor{black}{Ablation study on Text Encoder and Text Embedding Layer}}
\label{table4-4}
\centering
\resizebox{\textwidth}{!}{%
\tiny
\begin{tabular}{lcccccccc}
\toprule
\multirow{2}{*}{\textcolor{black}{\textbf{Method}}} &\multirow{2}{*}{\textcolor{black}{Text Encoding}} &\multirow{2}{*}{\textcolor{black}{Text Format}} & \multirow{2}{*}{\textcolor{black}{Param(M)}} & \multirow{2}{*}{\textcolor{black}{Flops(G)}} & \multicolumn{2}{c}{\textcolor{black}{\textbf{QaTa-COV19}}} & \multicolumn{2}{c}{\textcolor{black}{\textbf{MosMedData+}}} \\ \cmidrule(r){6-7} \cmidrule(l){8-9}
&           &                &           &                & \textcolor{black}{Dice (\%)}     & \textcolor{black}{mIoU (\%)}    & \textcolor{black}{Dice (\%)}    & \textcolor{black}{mIoU (\%)}    \\ \cmidrule{1-9}
\textcolor{black}{LViT-T} & \textcolor{black}{Text Encoder} & \textcolor{black}{Structured} & \textcolor{black}{84.9} & \textcolor{black}{101.8} 
  & \textcolor{black}{83.34} &  \textcolor{black}{74.76} &  \textcolor{black}{74.29} &  \textcolor{black}{61.18} \\
\textcolor{black}{LViT-T}  & \textcolor{black}{Embedding Layer} &  \textcolor{black}{Structured}  &  \textcolor{black}{29.7}  &  \textcolor{black}{54.1}   & \textcolor{black}{83.66}  & \textcolor{black}{75.11} & \textcolor{black}{74.57} & \textcolor{black}{61.33} \\
\cmidrule{1-9}
\textcolor{black}{LViT-T} & \textcolor{black}{Text Encoder} & \textcolor{black}{Unstructured} & \textcolor{black}{84.9} & \textcolor{black}{101.8} 
  & \textcolor{black}{82.53} &  \textcolor{black}{73.09} &  \textcolor{black}{73.71} &  \textcolor{black}{61.09} \\
\textcolor{black}{LViT-T}  & \textcolor{black}{Embedding Layer} &  \textcolor{black}{Unstructured}  &  \textcolor{black}{29.7}  &  \textcolor{black}{54.1}   & \textcolor{black}{82.41}  & \textcolor{black}{72.92} & \textcolor{black}{73.68} & \textcolor{black}{61.08} \\
 \midrule
 \vspace{-4mm}
\end{tabular}}
\end{table*}

\subsection{Ablation Study}
A series of ablation experiments are conducted to verify the performance of our LViT model, which is explored in the following four aspects. 

\subsubsection{Effectiveness of Proposed Components}
We perform relevant ablation experiments on the effectiveness of both supervised components and semi-supervised components of our LViT model, and the relevant experimental results are presented in Table \ref{table4-2}. In full supervision, we explore the effectiveness of these four components, i.e., DownViT, UpViT, PLAM, and Text. The Text refers to the text information. 
\textcolor{black}{And in the absence of text information, we utilize image features as the input of the transformer path.}
Experimental results illustrate that all of these components are effective, and the performance improvement brought by Text is the most significant.
\textcolor{black}{
Furthermore, we also conduct ablation experiments on two attention modules (GAP and GMP) in PLAM. Our findings suggest that GAP/PLAM outperforms GMP/PLAM, possibly because GAP is better at integrating  diverse information through global average pooling. It is worth noting that combining GAP and GMP yields better results than using them alone.}
To demonstrate our innovative points in semi-supervision, we explore the effectiveness of these three components, i.e., EPI, Text, and $L_{LV}$. 
\textcolor{black}{From the experimental results, it can be seen that the improvements of EPI, Text, and $L_{LV}$ are significant.} Among them, by incorporating text annotation information, the Dice score and mIoU score are improved by 1.59\% and by 1.66\% respectively when using 1/4 of the train set labels. And by introducing EPI mechanism, the semi-supervised performance of LViT is guaranteed to be comparable to the fully-supervised performance of U-Net. \textcolor{black}{LViT (1/4) with EPI yields a 0.26\% increase in the Dice score with text annotations and a 0.21\% increase in the Dice score without text annotations.} Finally, continuous improvements of model performance are also ensured by introducing $L_{LV}$ on unlabeled data. 
\textcolor{black}{
On the other hand, for labeled data, we already have the accurate mask for supervised learning. Therefore, using $L_{LV}$ on labeled data does not have significant benefits, as presented in the last row of Table \ref{table4-2}.}
\vspace{-2mm}
\begin{table}[!ht]
\caption{Ablation study with LViT-T on different Hyper-Parameters: Batch Size and Learning Rate.}
\label{table4-5}
\centering
\resizebox{\columnwidth}{!}{%
\large
\begin{tabular}{lccccc}
\toprule
\multicolumn{2}{l}{\multirow{2}{*}{\textbf{Hyper-Parameters}}}                                                   & \multicolumn{2}{c}{\textbf{QaTa-COV19}} & \multicolumn{2}{c}{\textbf{MosMedData+}}     \\ \cmidrule(r){3-4} \cmidrule(l){5-6}
\multicolumn{2}{l}{}                                                                                      & Dice (\%)          & mIoU (\%)          & Dice (\%)      & mIoU (\%)      \\ \cmidrule{1-6}
\multirow{3}{*}{\begin{tabular}[c]{@{}l@{}}Batch Size \end{tabular}}     & 16  & 82.72              & 73.96              & 73.98  & 61.10 \\
                      & 20  & 82.83              & 74.02              & 74.34 & 61.21 \\
                             & 24  & \textbf{83.66}     & \textbf{75.11}     &   \textbf{74.57}     & \textbf{61.33}  \\\cmidrule{1-6}  
\multirow{3}{*}{\begin{tabular}[c]{@{}l@{}}Learning Rate \end{tabular}} & 3e-4   & \textbf{83.66}     & \textbf{75.11}     &   74.52 & 61.31  \\
& 1e-3   & 82.25             & 73.69     & \textbf{74.57} & \textbf{61.33} \\
& 3e-3   & 82.20             & 73.53     &  74.46  & 61.27 \\\bottomrule       
\end{tabular}
}
\vspace{-2mm}
\end{table}

\subsubsection{Ablation Study on Model Size}
We conduct ablation experiments three times for model sizes to investigate the specific performance of LViT with different model sizes. Experiments are conducted on two datasets, QaTa-COV19 and MosMedData+, with six different model sizes, namely, LViT-TW/LViT-T, LViT-SW/LViT-S, LViT-BW/LViT-B, where "W" refers to without the text annotation, "T" refers to the tiny model, "S" refers to the small model, and "B" refers to the base model. 
\textcolor{black}{In the original vision transformer (ViT)\cite{42dosovitskiy2020image}, each ViT module has $L=12$ transformer layers. And the differences between different versions of LViT are in the number of Transformer layers in the DownViT module and UpViT module, where LViT-TW/LViT-T has only 1 Transformer layer per ViT module, LViT-SW/LViT-S has 4 Transformer layers per ViT module, and LViT-BW/LViT-B has 6 Transformer layers per ViT module.} Experimental results are reported in Table \ref{table4-4}.
As observed in the table, it is worth noting that LViT with the text annotation has only 1.7M more parameters and 0.1G more computation than LViT-W, while the improvement of segmentation performance brought by the text information is significant. 
\textcolor{black}{
Besides, there is also an interesting observation that larger models do not consistently improve accuracy over small models. We believe that it is related to the data distribution of the dataset. When the dataset distribution is relatively uniform, and the image is easy to segment, increasing the size of the model may not bring about a consistent improvement in performance. Conversely, if the dataset distribution presents notable differences and image segmentation is challenging, increasing the model size can lead to performance improvements. Nevertheless, it is noteworthy that as the model size increases, the model's performance jitter reduces, indicating the model becomes more robust.}

\subsubsection{Ablation Study on Hyper-Parameters}
Ablation experiments are conducted on two aspects of hyper-parameters: Batch Size and Learning Rate. For Batch Size, we set 16, 20, and 24 on the QaTa-COV19 dataset and the MosMedData+ dataset. According to Table \ref{table4-5}, LViT is optimal for batch size of 24 and learning rate of 3e-4 on the QaTa-COV19 dataset. And LViT is optimal for batch size of 24 and learning rate of 1e-3 on the MosMedData+ dataset. It is worth noting that the impact of hyper-parameters on the model performance is more considerable than the model size.

\subsubsection{Ablation Study on Text Encoder and Embedding Layer}
\textcolor{black}{
To further explore the encoding capabilities of text encoders and text embedding layers for text, as well as the differences in their applications to multimodal features, we conduct two sets of experiments for analysis. One set focuses on existing well-structured texts, while the other set focuses on poorly structured texts. We construct unstructured text annotations by randomly swapping the positions of phrases within the text. For example, we modify the description "Bilateral pulmonary infection, two infected areas, all left lung and middle lower right lung" to "All left lung and middle lower right lung, two infected areas, bilateral pulmonary infection". The experimental results, presented in Table V, reveal that utilizing a text encoder requires nearly three times as many parameters and nearly twice as much computation compared to using a text embedding layer. However, despite the increased complexity, the model performance does not improve and even decreases in well-structured reports. This finding supports our decision to employ a text embedding layer in our LViT model.  It is worth noting that the model performance with text embedding layer is slightly lower than  that of a text encoder for poorly structured reports. 
We believe this discrepancy can be attributed to the better encoding ability and robustness of text encoders when dealing with the more diverse radiology reports. However, it is important to acknowledge that the resulting parameter and computational costs are not cost-effective.}

\subsubsection{Ablation Study on Semi-Supervision}

\textcolor{black}{
Multiple semi-supervised experiments are conducted to verify the model performance in semi-supervised learning. These experiments cover two different label ratios, i.e., 25\% and 50\%, to explore the performance changes under different label ratios. Additionally, experiments are conducted with and without text information. We compare our method with both the traditional SOTA semi-supervised medical image segmentation methods, such as DTC \cite{luo2021semi}, PLCT \cite{chaitanya2023local}, and MC-Net+ \cite{wu2022mutual}, and the multimodal methods, such as LAVT \cite{Yang22CVPR} and GLoRIA \cite{huang2021gloria}.  The experimental results are presented in Table \uppercase\expandafter{\romannumeral6}, which demonstrate that our proposed LViT model achieves superior segmentation performance to other methods. This is attributed to the Exponential Pseudo-label Iteration mechanism and LV loss, regardless of whether text information is included in the pipeline or not.
}

\begin{table}[!ht]
\caption{Ablation study on semi-supervision with LViT-T and other methods on the QaTa-COV19 dataset}
\label{tableA-1}
\centering
\resizebox{\columnwidth}{!}{
\small
\begin{tabular}{ccccc}
\toprule
\textbf{Method} & Text & Label ratio & \textbf{Dice (\%)} & \textbf{mIoU (\%)} \\ 
\midrule
DTC \cite{luo2021semi} & $\times$ & 25\% & 76.07 & 66.04 \\
PLCT \cite{chaitanya2023local} & $\times$ & 25\% & 76.65 & 66.71 \\
MC-Net+ \cite{wu2022mutual} & $\times$ & 25\% & 76.93 & 67.02 \\
LViT-TW (1/4) & $\times$ & 25\% & \textbf{79.08} & \textbf{69.42} \\
\midrule
LAVT \cite{Yang22CVPR} & $\checkmark$ & 25\% & 77.08 & 67.21 \\
GLoRIA \cite{huang2021gloria} & $\checkmark$ & 25\% & 77.32 & 67.48 \\
LViT-T (1/4) & $\checkmark$ & 25\% & \textbf{80.95} & \textbf{71.31} \\ 
\midrule
DTC \cite{luo2021semi} & $\times$ & 50\% & 77.23 & 67.42 \\
PLCT \cite{chaitanya2023local} & $\times$ & 50\% & 77.66 & 68.04 \\
MC-Net+ \cite{wu2022mutual} & $\times$ & 50\% & 77.91 & 68.47 \\
LViT-TW (1/2) & $\times$ & 50\% & \textbf{80.35} & \textbf{70.74} \\
\midrule
LAVT \cite{Yang22CVPR} & $\checkmark$ & 50\% & 77.96 & 68.53 \\
GLoRIA \cite{huang2021gloria} & $\checkmark$ & 50\% & 78.49 & 68.97 \\
LViT-T (1/2) & $\checkmark$ & 50\% & \textbf{82.73} & \textbf{73.99} \\
\bottomrule 
\end{tabular}
}
\vspace{-2mm}
\end{table}

\begin{table}[!ht]
\large
\caption{Performance comparison between our method (LViT) and other methods on the ESO-CT dataset.}
\label{tableA-1}
\centering
\resizebox{\columnwidth}{!}{%
\begin{tabular}{ccccc}
\toprule
\textbf{Method} & Param (M) & Flops (G)& \textbf{Dice (\%)} & \textbf{mIoU (\%)} \\ \midrule
U-Net  & 14.8 & 50.3  & 66.75 & 56.31 \\
\textcolor{black}{U-Net++}  & \textcolor{black}{74.5} & \textcolor{black}{94.6}  & \textcolor{black}{66.70} & \textcolor{black}{56.59} \\
nnUNet & 19.1 & 412.7 & 68.38 &  56.10  \\
\textcolor{black}{TransUNet}  & \textcolor{black}{105} & \textcolor{black}{56.7}  & \textcolor{black}{65.94} & \textcolor{black}{55.78} \\
\midrule
LViT-TW  & 28.0  & 54.0 & 68.27  &  57.02  \\
LViT-T  & 29.7 & 54.1 & \textbf{71.53} & \textbf{59.94} \\ \bottomrule 
\end{tabular}
}
\vspace{-2mm}
\end{table}

\begin{figure*}[!ht]
\centering
  \includegraphics[width=\textwidth]{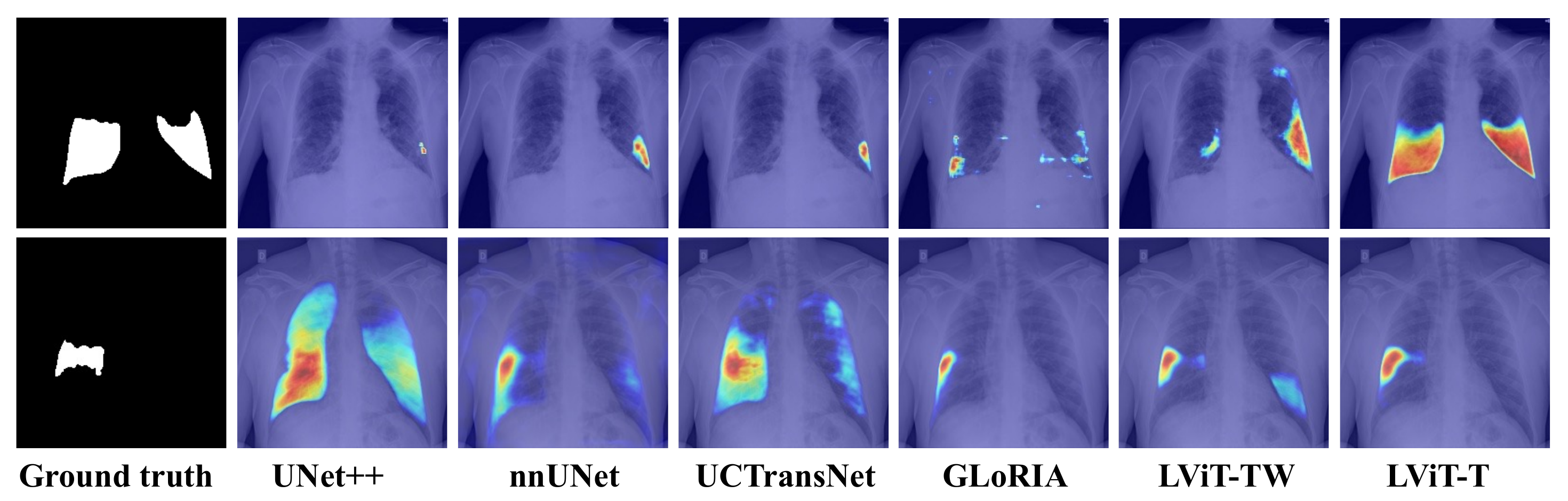}
  \caption{\textcolor{black}{Saliency map for interpretability study of different approaches on the QaTa-COV19 dataset. The language input of the first row is "Bilateral pulmonary infection, two infected areas, lower left lung and lower right lung". The language input of the second row is "Unilateral pulmonary infection, one infected area, middle left lung".}}
  \label{Inter1}
\end{figure*}

\begin{figure*}[!ht]
\centering
  \includegraphics[width=\textwidth]{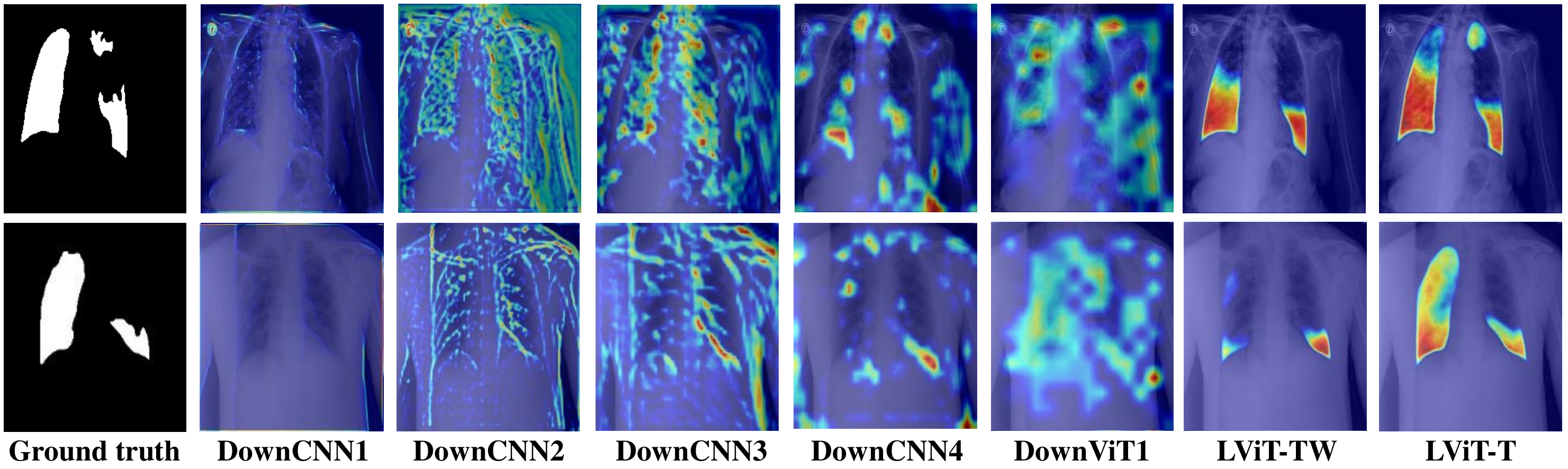}
  \caption{\textcolor{black}{Saliency map for interpretability study of different layers of LViT on the QaTa-COV19 dataset. The language input of the first row is "Bilateral pulmonary infection, three infected areas, all left lung and upper lower right lung". The language input of the second row is "Bilateral pulmonary infection, two infected areas, all left lung and lower right lung".}}
  \label{Inter2}
  \vspace{-2mm}
\end{figure*}

\subsection{Practical Application and Generalization Study on Esophageal CT Dataset}
\textcolor{black}{
To explore the generalization performance of LViT and demonstrate how text annotations can help in practical scenarios, we conduct experiments on the ESO-CT dataset.} This dataset is an esophageal cancer segmentation dataset collected by us, which consists of 286 cases. Each case contains mask and clinical information manually annotated by radiologist. In clinical information, the radiologist will divide the esophagus into four sections (Cervical, Upper Thorax, Middle Thorax, Lower Thorax) from top to bottom, and then give the rough location of tumor. The rough location of tumor is utilized as the text input in LViT. 
\textcolor{black}{
We compare our LViT model with several SOTA methods, including UNet, UNet++, nnUNet, and TransUNet.
As presented in Table. \ref{tableA-1}, the performance of LViT-TW and nnUNet is comparable and superior to other methods. It should be noted that nnUNet has an extremely complex preprocessing for images, while LViT does not need it. In addition, with text innoations, LViT-T outperforms other methods by a large margin,  thereby verifying the effectiveness and extensibility of LViT across different datasets.}
\textcolor{black}{
Therefore, we believe that with the approximate location (text annotations), the LViT model is able to segment the lesion area to form a more precise radiotherapy target area to better kill cancer cells.}

\subsection{Interpretability Study}
\label{sectionIS}
The interpretability study is performed on the QaTa-COV19 dataset to explore whether the LViT network can notice lesion regions and whether the introduction of text information can enhance the attention to lesion regions. 
\textcolor{black}{
In order to provide a more intuitive display of the changes in the region of interest of the model, we utilize GradCAM \cite{selvaraju2017grad} to compare the activation for regions of attention.}
Figure \ref{Inter1} shows that UNet++, nnUNet, UCTransNet, and GLoRIA all have different degrees of misactivation regions. For example, in the case of total lung infection, these methods can only activate half of the lung region. In contrast, by introducing text information, our model can activate more regions, and the edge profile of the activated regions of interest is more consistent with the ground truth. Therefore, the localization of lesions can be learned by the text input. 

\textcolor{black}{
Besides, to better explore the benefits of incorporating text information into the pipeline, we conduct more experiments on another two cases with more segmentation areas, as shown in Figure \ref{Inter2}.}
We perform activation mapping in DownCNN1, DownCNN2, DownCNN3, DownCNN4, and DownViT1, respectively, where DownCNN1 and DownViT1 represent the first layer DownCNN and the first layer DownViT, respectively. The text information is input to the model through DownViT1, thus the difference in activation regions between DownViT1 and DownCNN1 can be approximated as the difference brought by the text information. It can be seen that the activation effect of the region of interest of DownViT1 is similar to that of DownCNN4. It is also worth noting that image feature of DownViT1 comes from DownCNN1, which failed to activate the lesion region but only activated the lung boundary. However, DownViT1 can directly activate the relevant lesion region by introducing the text information. And it indicates that the text information can effectively help locate lesion region in the lung, thus prompting the network to pay more attentions on the region indicated by the text information. \textcolor{black}{The CAM output of LViT-TW and LViT-T shows the final activation difference caused by the text information. By comparing the regions of interest for LViT-TW/LViT-T, we believe that the text information can help reduce the probability of mis-segmentation.}

\section{Conclusion}
In this paper, we propose a new vision-language medical image segmentation model LViT, which leverages medical text annotation to compensate for the quality deficiency in image data and guide to generate pseudo labels of improved quality in the semi-supervised learning. Multimodal medical segmentation datasets (image + text) are constructed to evaluate the performance of LViT, and experimental results show that our model has superior segmentation performance in both fully-supervised and semi-supervised settings.
\textcolor{black}{In addition, we present an example application on the diagnosis and treatment of early-stage esophageal cancer to demonstrate how text annotations can help in practical scenarios.}
Currently, the proposed model is a 2D Segmentation model. In our future work, we will extend our model to 3D and conduct experiments on more medical data to further verify its generality.
\textcolor{black}{Besides, in the current version of our LViT model, it is necessary to supply text inputs during the inference stage. Therefore, our another future work is to generate text annotation automatically according to the provided image information.} As the text annotations are structured, we can transform the problem of text annotation generation into a classification problem in the future version of LViT. This will enable us support inference either with or without text input, thereby enhancing the usability of our model.

\bibliographystyle{ieeetr}
\bibliography{main}

\end{document}